\pdfoutput=1

\documentclass[11pt]{article}

\usepackage{ACL2023}

\usepackage{times}
\usepackage{latexsym}

\usepackage[T1]{fontenc}

\usepackage[utf8]{inputenc}

\usepackage{microtype}

\usepackage{xcolor}
\usepackage{tabularx}

\usepackage{amsfonts}       
\usepackage{amsmath}
\usepackage{amsthm}
\usepackage{amssymb}
\usepackage{nicefrac}       
\usepackage{xfrac}
\usepackage{subcaption}
\usepackage{multirow}
\usepackage{ctable} 
\usepackage{booktabs}       
\usepackage{diagbox}
\usepackage{csquotes}
\usepackage{enumitem}
\usepackage{inconsolata}
\usepackage{lipsum}
\usepackage{abraces}

\title{$\mathtt{HumBEL}$: A Human-in-the-Loop Approach for Evaluating Demographic Factors of Language Models in Human-Machine Conversations}

\author{Anthony Sicilia$^\flat$ \quad Jennifer C. Gates \quad Malihe Alikhani$^\flat$ \\
$^\flat$Northeastern University \\
\texttt{\{sicilia.a, m.alikhani\}@northeastern.edu} \quad \texttt{gatesjenniferc@gmail.com}}

\begin{document}
\maketitle
\begin{abstract}
While demographic factors like age and gender change the way people talk, and in particular, the way people talk to machines, there is little investigation into how large pre-trained language models (LMs) can adapt to these changes. To remedy this gap, we consider how demographic factors in LM language skills can be measured to determine compatibility with a target demographic. We suggest clinical techniques from Speech Language Pathology, which has norms for acquisition of language skills in humans. We conduct evaluation with a domain expert (i.e., a clinically licensed speech language pathologist), and also propose automated techniques to complement clinical evaluation at scale. Empirically, we focus on age, finding LM capability varies widely depending on task: GPT-3.5 mimics the ability of humans ranging from age 6-15 at tasks requiring inference, and simultaneously, outperforms a typical 21 year old at memorization. GPT-3.5 also has trouble with social language use, exhibiting less than 50\% of the tested pragmatic skills. Findings affirm the importance of considering demographic alignment and conversational goals when using LMs as public-facing tools. Code, data, and a package will be available.
\end{abstract}
\section{Introduction} 
\label{sec:intro}
\begin{figure}[t]
    \centering
    \includegraphics[width=\linewidth]{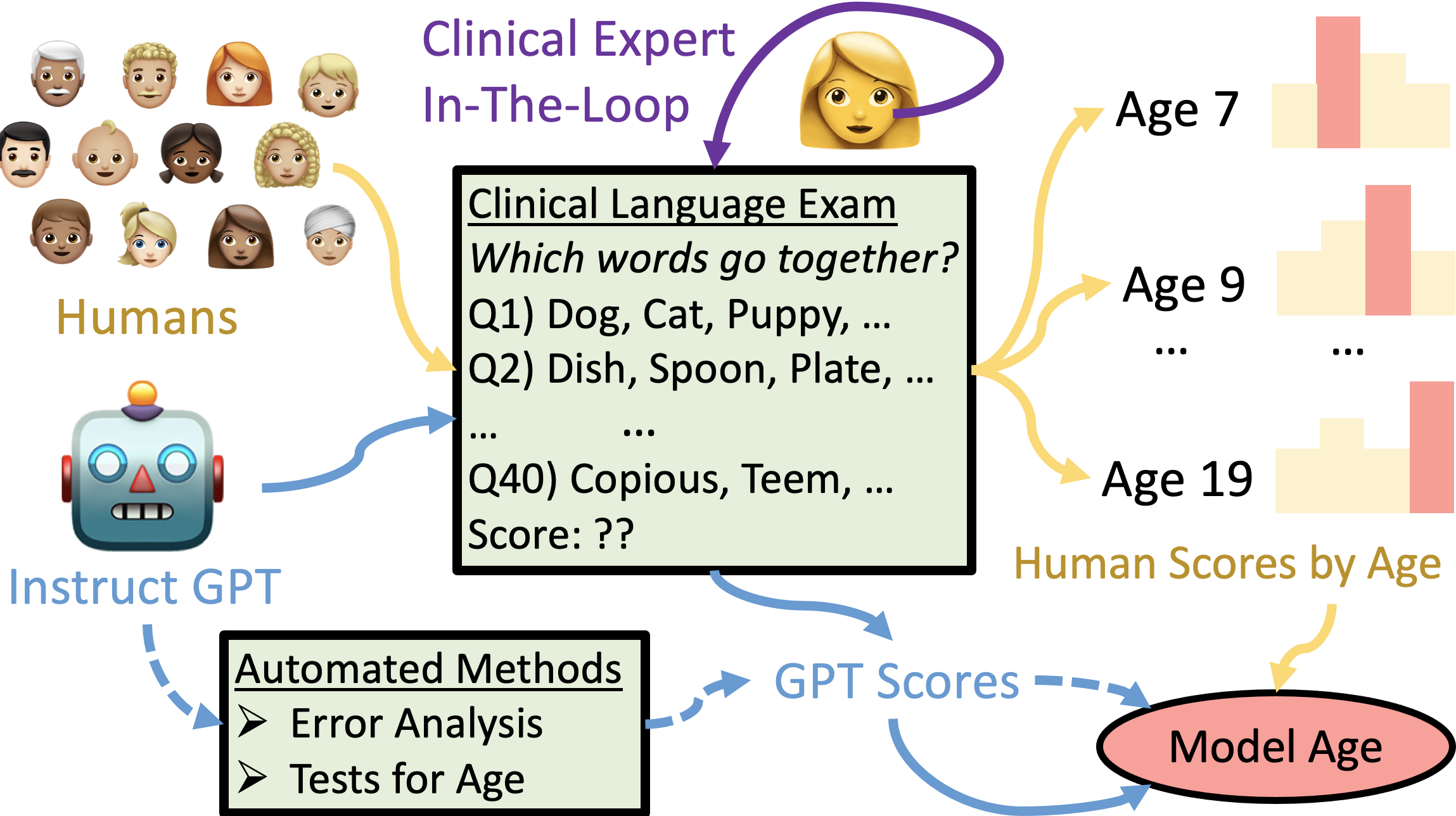}
    \caption{\small $\mathtt{HumBEL}$ uses data from human clinical exams to measure demographic factors of language models (LMs) and test alignment of LM language use with demographic groups. We propose human-in-the-loop and automated techniques.}
    \label{fig:intro}
\end{figure}
Demographic factors like age and gender impact the words we use \citep{sap-etal-2014-developing, giorgi2021characterizing} and, more broadly, the way we interact and communicate with each other \citep{de2022social}. Moreover, these same factors carry over influence into our conversations with machines. Age group, in particular, impacts the way we converse with household dialogue systems like Alexa \citep{pradhan2019phantom}, conversational agents for health information access \citep{harrington2022s}, and intelligent systems for interactive tutoring \citep{ogan2012oh}. Ultimately, to effectively communicate, dialogue systems must adapt and align with the pragmatic skills, semantic understanding, and common sense of their target demographic. Despite this, there is limited work on evaluating demographic factors, and in particular, demographic alignment in human-machine conversations. To fill this gap,
we propose the novel $\mathtt{HumBEL}$ evaluation framework,\footnote{\underline{$\mathtt{Hum}$}an demographic \underline{$\mathtt{B}$}ased \underline{$\mathtt{E}$}valuation of \underline{$\mathtt{L}$}Ms} which measures demographic alignment of language models (LMs) with a target user demographic for the first time. 
While our framework is general, we pay particular attention to modern LMs to support the rapid development of these technologies as public-facing tools. 

In detail, $\mathtt{HumBEL}$ proposes a human-in-the-loop evaluation protocol which collaborates with a field of clinical experts (Speech Language Pathologists) that have already actively studied demographic factors in human-human communication for over 98 years \cite{duchan2023charter}. These clinical experts administer language exams and compare to normative data (from large, human patient populations) to determine whether a patient aligns with a target demographic (e.g., their peers).
$\mathtt{HumBEL}$ works by collaborating with these domain-experts to administer these same tests to a language model (LM), so key differences between LMs and human sub-populations are revealed (Figure~\ref{fig:intro}). To complement our human-in-the-loop clinical exams, we also propose a novel statistical test and a suite of existing statistical techniques to confirm clinician findings at scale. While $\mathtt{HumBEL}$ is generally applicable to any (categorical) demographic features, we focus this study on age demographics. Most importantly, our evaluation of LM alignment with different age categories can be used to examine robustness in matching conversation applications, but as a side-effect, our techniques are also able to assign a typical human age-equivalent to an LM for a specific language skill.\footnote{See~\hyperref[sec:limits]{\textcolor{black}{\textbf{Limitations}}}. Significant care should be taken in interpretation of LM age equivalents.}

To demonstrate $\mathtt{HumBEL}$, we evaluate OpenAI's suite of GPT-3.5 models. Our key findings quantify gaps in commonsense knowledge (about noun relationships), social language use, and inference skills compared to adult human populations. Further, we find inconsistency in language skills compared to normal human development: failures in social and inferential capability are akin to error patterns of a typical 3 year old at worst or 15 year old at best, while success at recollection surpasses a typical 21 year old. Results highlight the potential for human-machine miscommunication, when the demographic factors of conversation are ignored. 

Hereinafter, we introduce $\mathtt{HumBEL}$, contributing: 
\begin{enumerate}[leftmargin=*, nolistsep]
\item (\S~\ref{sec:clinical}) protocols for evaluation of demographic factors in LMs by domain experts, using clinical exams 
and detailed clinician error analyses;
\item (\S~\ref{sec:automation}) statistical tools to complement clinical techniques at scale via 
novel statistical tests for demographic alignment
and error analysis;
\item (\S~\ref{sec:results}) comprehensive evaluation of a modern LM (GPT-3.5) using our aforementioned techniques;
\item (\S~\ref{sec:chat-compare}) comparison of GPT-3.5 with other modern LMs, using our statistical tools;
\item and code, data, and a python package for future researchers to easily apply our tests.\footnote{Resources at: \href{https://github.com/anthonysicilia/humbel}{https://github.com/anthonysicilia/humbel}}
\end{enumerate}
%
\section{The $\mathtt{HumBEL}$ Framework: Human Age Based Evaluation of Language Models}
\label{sec:humbel}
$\mathtt{HumBEL}$ consists of two evaluation protocols. The first (preferred) protocol describes techniques to administer a clinical exam to a LM via prompting, so that results can be carefully analyzed by a clinically licensed Speech Language Pathologist. The second describes automated alternatives, which are easier to conduct more frequently and at scale. 
\subsection{Clinical Evaluation by Speech Language Pathologist}
\label{sec:clinical}
In this section, we use examples from the commonly used CELF5 clinical exam \citep{wiig2003celf} to describe our protocols.\footnote{\label{TOU} Note, any examples of test materials provided during discussion are \textit{adaptions} of the original materials per publishing agreement with Pearson, Inc. While different, the examples are designed to convey similar qualitative insight to the reader; e.g., the LM prompt or types of errors made by the LM.}$^,$\footnote{\textit{Clinical Evaluation of Language Fundamentals, Fifth Edition, CELF-5} Copyright © 2013 NCS Pearson, Inc. Reproduced with permission. All rights reserved.}$^,$\footnote{\textit{Clinical Evaluation of Language Fundamentals, Fifth Edition, CELF-5} is a trademark, in the US and/or other countries, of Pearson Education, Inc. or its affiliates(s).}  This test is used throughout our paper, but our ideas generalize to other common clinical tests.
\subsubsection{Description of CELF5 Exam}
\label{sec:clinical_descs}
CELF5 is composed of multiple sub-tests with 24-50 questions each. We consider the sub-tests below, which are designed to assess syntactic, semantic, and pragmatic use of language in 5-21 year olds.
\begin{enumerate}[leftmargin=*, nolistsep]
    \item \textbf{Word Classes (WC)} presents 3-4 words and asks test subject to identify the two words that go together best. It measures semantic knowledge and ability to apply this knowledge to determine and rank word associations. 
    \item \textbf{Formulated Sentences (FS)} presents 1-2 words and asks subject to provide a sentence which uses the(se) word(s). It measures syntactic and semantic correctness of the provided sentence. 
    \item \textbf{Recalling Sentences (RS)} presents a sentence and asks subject to repeat the sentence. It measures short-term memory and reproduction skill. 
    \item \textbf{Understanding Spoken Paragraphs (USP)} presents a story and asks subject questions about the story. It primarily measures recollection ability with occasional need for inference. 
    \item \textbf{Pragmatics Profile (PP)} analyzes social error patterns of subject, observed throughout other sub-tests as well as more targeted interactions. 
\end{enumerate}
\begin{table*}[]
    \centering \small
    \begin{tabularx}{\textwidth}{X|X|X}
    \texttt{SLP} & \texttt{QA} & \texttt{Comp} \\\hline
     Carefully consider the following words and tell me the two words that go together best: "[W]", ... & Instruction: Carefully consider the following words and tell me... 

     Student: & Among the words "[W]", "[X]", "[Y]", and "[Z]", the two words that go together best are
    \end{tabularx}
    \caption{\small Examples from different prompt protocols for the Word Classes test. \texttt{SLP} follows CELF5 directives exactly.$^\text{\ref{TOU}}$ \texttt{QA} adds a mechanism to inform the LM of its speaker role. \texttt{Comp} re-frames as a likely seen prefix (i.e., in training). We test these and 70+ other prompt/parameter variations. See sensitivity analysis in Appendix~\ref{sec:sensi_tests}.}
    \label{tab:prompts}
\end{table*}
\subsubsection{Exam Administration via Prompting} Prompting is the standard technique in which textual output is generated from LMs. We use \textit{prefix prompting}, in which input text is provided to the LM and the LM is sampled based on this input to complete the text. In this way, questions from the 5 discussed tests can be administered to the LM and the LM response (i.e., the text-completion) can be evaluated by the clinician with relevant observations noted for each question. Since the integrity of exam results requires precise adherence to the CELF5 protocols for scoring/evaluation, we adhere to these as much as possible. We do identify two primary limitations in administering CELF5 to common LMs and provide solutions below:
\begin{enumerate}[leftmargin=*, nolistsep]
    \item First, some \textit{LMs are better suited for text-completion than instruction following}, making typical administration of the test challenging. To control for performance drops induced by this, we use multiple prompt formats (see Table~\ref{tab:prompts}). The \texttt{SLP} protocol follows the CELF5 directives exactly, while the \texttt{QA} and \texttt{Comp} protocols should be better tailored for LMs. Sensitivity analysis (Appendix~\ref{sec:sensi_tests}) with 70+ additional configurations suggests prompt and parameter variations do not significantly impact LM performance.
    \item Secondly, \textit{LMs lack the ability to perceive visually and take action in an embodied setting}. Therefore, we limit the types of tests administered (i.e., those in \S~\ref{sec:clinical_descs}) and tailor these tests for a language-only medium when appropriate (see \hyperref[sec:mods]{\textcolor{black}{\textbf{Modifications}}}). Investigation of the impact of this choice is left for future work. Indeed, the necessity of visual/embodied stimuli to inform lexical semantics has been hypothesized \cite{Bisk2020} and CELF5 scores may be used in the future to provide a principled answer. 
\end{enumerate}
\subsubsection{Exam Administration via Chat}
While the experimental focus is on text-completion prompts, we also conduct analysis of chat-based models, like ChatGPT. Here, we can follow CELF5 directives more precisely, but still modify tests to accommodate the limited turn-based chat medium; i.e., removing visual cues, taking scores with/without evaluation of non-verbal skills, etc.
\subsection{Automation of Clinical Techniques}
\label{sec:automation}
In this part, we describe automated techniques for two important aspects of the clinical exam: (1) qualitative analysis of errors through clinician notes and (2) determination of human demographic alignment for the LM on a task. We use the \textbf{Word Classes} test (\textbf{WC}) as an example application.
\subsubsection{Data}
\label{sec:wc_large}
We build a large-scale \textbf{WC} test (\textbf{WC} \texttt{large}) by combining two publicly available data sources:
\begin{enumerate}[leftmargin=*, nolistsep]
    \item \textbf{Word Associations}: We build associated word pairs using \textit{cue} and \textit{association} words from the WAX dataset \citep{liu-etal-2022-wax} collected from human annotators by presenting a \textit{cue} and asking for spontaneous associations (with explanation). 
    This dataset is transformed into a large-scale version of the \textbf{WC} test by randomly sampling two additional association words for each human labeled word pair and presenting the quadruple to a subject using the existing \textbf{WC} prompt protocols.
    All four test words (i.e., the target pair and two additional associations) are presented in random order and filtered to prevent overlap in target pairs by chance.
    \item \textbf{Age Norms}: 
    In clinical exams, human developmental standards are determined from exam score data (i.e., \textit{age norms}) that indicate the age at which one expects the observed score in a human population. To do this automatically for new \textbf{WC} questions, we use a test-based \textit{age-of-acquisition} (AoA) dataset \citep{dale1976living, brysbaert2017test}, which determines the AoA of 40K English words. Word AoA is determined by the age at which 50-70\% of a human population \textit{knows} the word according to a definition matching test (see Appendix~\ref{sec:word_aoa}), called \textbf{Def} in experiments (\S~\ref{sec:results}). For \textbf{WC} \texttt{large}, AoA is the max AoA of the target words (i.e., the typical age at which a human can select the target pair without guessing).
\end{enumerate}
Applying AoA estimates to the word association data leads to about 10K new \textbf{WC} questions with accompanying explanations and projected age norms. 
\subsubsection{Automated Analysis of Errors}
\label{sec:error_analysis}
We isolate some influential factors in typical word acquisition by humans based on discussion with a licensed Speech Language Pathologist; i.e., these question/response features were deemed useful for analyzing errors in notes during clinical exams. We limit our analysis to features that can be automatically determined.\footnote{We use the \href{https://spacy.io}{spacy} package.} The target pair features include: unordered \textbf{parts-of-speech} inferred from explanations in the WAX dataset, \textbf{relation types} from the WAX dataset, and \textbf{morphological complexity}. We also consider presence of \textbf{explanations} by GPT. Details on feature extraction are in Appendix~\ref{sec:feature_details}.
\paragraph{Statistical Tests} In lieu of detailed notes, we propose a variety of statistical tests to determine association and impact of the various features just discussed. The $\chi^2$-statistic provides a basic test for the association of each feature with the occurrence of an LM error. Furthermore, specific hypotheses about the impact of particular parts-of-speech, relations, and other features can be estimated using a Linear Probability Model (LPM). For example, an LPM allows us to estimate the effect size
\begin{equation}\small
\label{eqn:prob_inc}
\begin{split}
& \mathbf{Pr}\{\text{LM error} \mid \text{Relation=Function}\} \\
& \quad - \mathbf{Pr}\{\text{LM error} \mid \text{Relation}\ \neq \ \text{Function}\}.
\end{split}
\end{equation}
while controlling for other features such as typical human age-of-acquisition for the word pair and any other features included in the model. \textit{To summarize}, the $\chi^2$ test lets us test the basic association between the occurrence of errors and automatically determined features, whereas the LPM lets us directly test more complicated hypotheses, e.g., ``controling for AoA, does the chance of an error increase when the word pair has a functional relation?'' For details on both procedures see Appendix~\ref{sec:tools}. Example applications are provided in later results (\S~\ref{sec:results}).
\subsubsection{Automated Determination of LM Age}
\label{sec:lm_age}
While we focus on age, these novel statistical tests can measure any categorical demographics.
\paragraph{Test Divergence} We base our first test for LM age on a statistic called the \textit{test divergence} \citep{sicilia-alikhani-2022-leather}. For an evaluation function $h$ and language model \texttt{LM} the test-divergence is:
\begin{equation}
\label{eqn:testdiv}\small
\begin{split}
& \mathbf{TD}_a(\texttt{LM}) = \mathbf{E}[\lvert h(D) - h(\hat{D})\rvert]; \\ 
& \qquad (D, C) \sim \mathbb{G}_a; \quad \hat{D} \sim \texttt{LM}(C).
\end{split}
\end{equation}
Here, $\mathbb{G}_a$ is called the goal distribution and typically represents a distribution of human dialogues. We incorporate new dependence on the age group $a$, which restricts the human reference population. With this interpretation, $D$ is a random human dialogue about the context $C$ and $\hat{D}$ is a dialogue sampled from the language model about this same context; context can be a prompt, an image, both (for perceptually grounded models), or any other information source which grounds the dialogue. In this paper, $C$ will correspond to a test question (or, equivalent LM prompt) in the \textbf{WC} \texttt{large} dataset and $h$ will indicate whether the response $D$ (or $\hat{D}$) is correct. $C$ follows a uniform distribution over questions in \textbf{WC} \texttt{large} where AoA (\S~\ref{sec:wc_large}) is either (1) exactly equal to $a$, or (2) $\leq a$. We disambiguate between these two cases throughout.
\paragraph{The $\mathbf{TD}$ Test for LM Age} Granted the test-divergence as a test statistic, we are interested in the following null $H_0$ and alternative $H_A$ hypotheses:
\begin{displayquote}\small
$H_0:$ \texttt{LM} errors align with age group $a$ \\
$H_A:$ \texttt{LM} errors fail to align with age group $a$
\end{displayquote}
Thus, we grant the LM benefit of the doubt and reject the model \texttt{LM} aligns with an age group if we establish evidence against this claim. Formally, we define \textit{alignment} when a model's error patterns are within a tolerance $\gamma$: i.e.,  if $\mathbf{TD}_a(\texttt{LM}) \leq \gamma$. In English, this means the expected difference between the \texttt{LM} performance and human (aged $a$) performance on each test question is no more than the tolerance $\gamma$ where tolerance allows us to account for any (human) subjectivity in question responses.
Then, with this, we can rewrite our hypotheses:
\begin{displayquote}\small
$H_0: \mathbf{TD}_a(\texttt{LM}) \leq \gamma$, \
$H_A: \mathbf{TD}_a(\texttt{LM}) > \gamma$.
\end{displayquote}
In turn, a test at confidence $100 \times (1-\alpha) \%$ rejects the null if the $p$-value is bounded by $\alpha$
\begin{equation}\small
p = \mathbf{Pr}(\widehat{T}_a - \gamma \leq T_a - \gamma \mid H_0) \leq \alpha
\end{equation}
where $\widehat{T}_a$ is the observed estimate of $\mathbf{TD}_a(\texttt{LM})$ (i.e., an empirical average) and $T_a$ is the r.v. representing this empirical average. For the \textbf{WC} \texttt{large} dataset, $n \cdot T_a$ is a Binomial random variable and probability under the Binomial distribution gives the $p$-value exactly. In other cases, the test outcome may be continuous or the test $h$ may be learned from data similar to work by \citet{bruni-fernandez-2017-adversarial}. Here, Hoeffding's or PAC type bounds can yield $p$-values \citep{shalev2014understanding}. 
\paragraph{The Mean Test for LM Age} As we will see in later results, the statistic/test just described will often be preferred because it incorporates information about individual question outcomes, making it more sensitive to correlation between $h(D)$ and $h(\hat{D})$. Still, we may not have access to the individual human question outcomes $h(D)$. Instead, we might only know the average outcome $\mu_a = \mathbf{E}[h(D)]$ with $D \sim \mathbb{G}_a$. Following the same logic as before, we can use this to test alignment:
\begin{displayquote}\small
$H_0: \mathbf{E}[R] = n \cdot \mu_a$, \
$H_A: \mathbf{E}[R] < n \cdot \mu_a$.
\end{displayquote}
where $R$ is the empirical sum of correct GPT responses $\sum_i h(\hat{D}_i)$ and $n$ is the question count. Note, this leads to a standard Binomial test.
%
%
\section{Results: Applying $\mathtt{HumBEL}$ to GPT}
\label{sec:results}
\subsection{Clinical Evaluation Results}
\label{sec:clinical_results}
%
\begin{table*}[]
    \centering \small
    \begin{tabular}{
  p{\dimexpr.06\linewidth-2\tabcolsep-1.3333\arrayrulewidth}
  |p{\dimexpr.68\linewidth-2\tabcolsep-1.3333\arrayrulewidth}
  |p{\dimexpr.26\linewidth-2\tabcolsep-1.3333\arrayrulewidth}
  }
    Test & \texttt{InstructGPT} & Clinician Observation \\\hline
    \textbf{WC} & Among the words "car", "water", "stroller", and "boat", the two words that go together best are "car" and "boat". Both are types of transport. & Misses functional \textit{goes in} relation for \textit{boat}, \textit{water} \\ 
    \textbf{WC} & Among the words "singing", "loving", "touching", and "tasting" the two words that go together best are "singing" and "loving." This is because both words involve using your voice and express affection. &  Misses categorical \textit{sense} relation for \textit{touch}, \textit{taste} \\
    \textbf{USP} & Melanie greeted Miss Grace because she was happy to see her. & Missing context: Grace is old camp instructor \\
    \textbf{USP} & ["throw-and-chase" is] a game where one person throws a ball and the other person goes to chase it. & Not fact based. Fact-based answer is found in context
    \end{tabular}
    \caption{\small Examples of inferential and other language application errors by InstructGPT in CELF5 exam. Explanations are provided by the evaluating clinician. Examples are adapted for publication per agreement with Pearson.$^{\text{\ref{TOU}}}$}
    \label{tab:examples}
\end{table*}
\begin{table*}[]
    \centering \small
    \begin{tabular}{c|c|c|c|c|c|c|c||c|c|c|c|c}
        \textbf{model} & \textbf{WC} & \textbf{WC}$^*$ & \textbf{FS} & \textbf{RS} & \textbf{USP} & \textbf{PP} & \textbf{PP}* & \textbf{WC} & \textbf{WC}$^*$ & \textbf{FS} & \textbf{RS} & \textbf{PP} \\\hline
         \texttt{Instruct w/ SLP \ } &  3\% & 50\% & 94\% & 88\% & 93\% & & & 3:2 & 7:5 & 21:5+ & 21:5+ & \\ 
         \texttt{Instruct w/ QA \ \ } & 28\% & 50\% & 85\% & 96\% & 93\% & 39\% & 48\% & 5:3 & 7:5 & 12:7 & 21:5+ & < 3 \\
         \texttt{Instruct w/ Comp } & 35\% & 60\% & 90\% & 100\% & 88\% & & & 5:11 & 8:10 & 15:1 & 21:5+ & \\\hline
         \texttt{ChatGPT (0301)\ \ \  } & 83\% & 83\% & - & - & 75\% & 45\% & 60\% & 14:7 & 14:7 & - & - & < 3
    \end{tabular}
    \caption{\small Evaluation of GPT-3.5 (\texttt{text-davinchi-002}, \texttt{turbo-0301}). (Left) Test scores reported as percent of highest possible score. (Right) CELF5 age equivalent (year:month) for scores on Left. CELF5 age equivalents are not available for \textbf{USP} or \textbf{PP}$^*$. Discussion focus in \S~\ref{sec:results} is placed on InstructGPT, while ChatGPT and other models (Table~\ref{tab:many_models}) are discussed in \S~\ref{sec:chat-compare}.}
    \label{tab:scores}
\end{table*}
\begin{table}[]
\centering\small
\begin{tabular}{l|l|lll}
\textbf{model}    & \textbf{age range} & \multicolumn{3}{c}{\textbf{accuracy}} \\
& & \textbf{@7} & \textbf{@15} & \textbf{@19} \\\hline
Llama-2-chat 7B    & < 7 & 24.2    & 19.1     & 23.4     \\
Zephyr-$\beta$ 7B  & < 7 \ to \ 7           & 36.5    & 27.4     & 21.5     \\
Mistral 7B v0.2 & \ \ \ 7 \ \hspace{.3mm}to \ 11          & 58.7    & 39.6     & 33.8     \\ \hline
InstructGPT & < 7 \ \hspace{.1mm}to \ 9           & 48.8    & 42.3     & 45.0     \\
ChatGPT (0301) & \ 15  \ \hspace{.3mm}to \ 19+         & 71.5    & 65.2     & 62.4 \\
ChatGPT (1106) & \ 15  \ \hspace{.3mm}to \ 19+         & 66.7    & 62.5    & 66.2 \\
\quad + \ \ 3 examples & < 7 \ to \ \hspace{.1mm}13$^*$         & 52.4    & 37.5    & 55.4 \\
\quad + 10 examples & < 7 \ to \ \hspace{.1mm}19+$^*$         & 51.6    & 44.2    & 68.8 \\
\end{tabular}
\caption{\small Age estimates using Mean Test on \textbf{WC} \texttt{large} with different models. $\mu_a$ is estimated as in Figure~\ref{fig:large-scale-pvals}. Age range upper- and lowerbounds use \textit{smaller} and \textit{larger} estimates of $\mu_a$, respectively, providing a \textit{less} strict and \textit{more} strict test of age. For both, we report the first age $a$ at which significant difference is noted when AoA = $a$. Accuracy for different word AoA is also provided. Random sample of 1000 examples total are used for 7B models and ChatGPT 1106. For open-source 7B parameter models, quantization is used for inference. For ChatGPT 1106, we test impact of in-context learning using 3 or 10 random demonstrations; $^*$Mean Test is invalid for ICL.}
\label{tab:many_models}
\end{table}
%
%
\begin{figure}
    \centering
    \includegraphics[width=.9\linewidth]{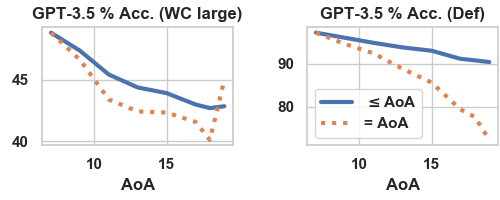}
    \caption{\small Accuracy of InstructGPT on \textbf{WC} \texttt{large} and \textbf{Def}.; AoA is defined in \S~\ref{sec:wc_large}. 
    Solid line tests pairs at most the AoA. Dotted tests pairs exactly at the AoA.
    }
    \label{fig:large-scale-acc}
\end{figure}
\begin{figure}
    \centering
    \includegraphics[width=.9\linewidth]{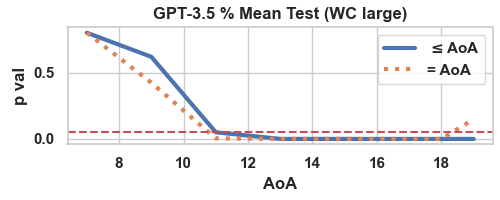}
    \caption{\small Vertical axis shows $p$-values from mean tests. Red dashed line is $\alpha = 0.05$. $\mu_a$ is estimated based on \citet{dale1976living}, accounting for chance and subjectivity of gold associations (see Appendix~\ref{sec:hum_mean_correct}).}
    \label{fig:large-scale-pvals}
\end{figure}
Table~\ref{tab:scores} shows CELF5 test scores and age equivalents for InstructGPT (\texttt{text-davinci-002}) and select results for ChatGPT (\texttt{gpt-3.5-turbo-0301}).\footnote{Previously accessible through the \href{https://platform.openai.com/docs/deprecations}{OpenAI API}.} We discuss qualitative clinician observations with supporting quantitative analyses, \textit{providing italicized takeaways for conversational applications}. While this part focuses on InstructGPT, comparison to ChatGPT is provided in \S~\ref{sec:chat-compare}. For sensitivity analysis to prompt/parameters, see Appendix~\ref{sec:sensi_tests}.
\paragraph{Modifications}\label{sec:mods}
To adapt the \textbf{Word Classes} for language models, we remove any visual stimuli. We also include a further modified test \textbf{WC}$^*$. While official clinical evaluation stipulates the evaluator should prematurely conclude the \textbf{WC} test if 4 sequential incorrect answers are provided, this stopping rule (ceiling) is based on human development (i.e., easier words are presented earlier), which GPT may not follow. For comparison, \textbf{WC}$^*$ reports evaluation without a ceiling. Similarly, we modify the \textbf{Pragmatics Profile} \textbf{PP} since it measures social language capabilities which are not observable in prompt-only or turn-based chat mediums; e.g., non-verbal cues and initiative behaviors. The profile with these items removed is called \textbf{PP}$^*$. 

%
\paragraph{Recollection vs. Inference} \textit{InstructGPT excels at memorization, but has trouble making inferences.} Of all tests in Table~\ref{tab:scores}, Word Classes (\textbf{WC}) most requires the ability to make new inferences from existing (lexical semantic) knowledge. This is also the task that InstructGPT performs worst at, demonstrating alignment with the ability of a 6 year old. While InstructGPT was generally more successful on other tasks, the evaluating clinician observed  errors in \textbf{USP} 
were also frequently due to trouble drawing inferences. 
When InstructGPT provided explanations for answers on \textbf{WC}, the clinician observed flawed or irrelevant logic in more than 59\% of cases. See Table~\ref{tab:examples} for examples of inferential and other language application errors. Note, this pitfall of GPT also induces a large variation in scores (e.g., from age equivalent over 21 to under 4) which is certainly atypical of human norms. Despite some negatives, the impressive proficiency of GPT at recollection suggests \textit{it would excel in conversational applications requiring rote information extraction}. In applications requiring inference about word meanings, \textit{one might consider communicating the error patterns of GPT, depending on target interlocutor age and conversational goals.} 
\paragraph{Difficult Relations} \textit{InstructGPT has more trouble with functional roles, categories, and antonyms.} On Word Classes (\textbf{WC}), the evaluating clinician identified multiple errors for each of these relation types. For functional roles, InstructGPT fails to recognize relationships like "[X] goes in [Y]" or "[X] used for [Y]". It also failed to recognize categories like "body parts", "senses" and dichotomous pairs used to describe the same concept; e.g., "brief" and "long".$^{\text{\ref{TOU}}}$ Table~\ref{tab:examples} shows examples.
\paragraph{Atypical Semantic Errors} \textit{According to human developmental standards, InstructGPT understands some "hard" words better than "easy" words.} In particular, the clinician observed error patterns in semantic knowledge which were distinct from typical patterns in children. While InstructGPT failed frequently at comparatively "easy" word relations (e.g., \textit{shirt} and \textit{jacket}), it succeeded at "harder" relations (e.g., \textit{copious} and \textit{teem}).$^{\text{\ref{TOU}}}$ In the data, this is exemplified by \textbf{WC} and the modified test \textbf{WC}$^*$. The difference in scores implies InstructGPT accumulated sequential errors early in the test on "easy" word relations, while still succeeding later on "hard" relations. This example hits home the necessity of considering human demographics in evaluation, since\textit{ GPT does not appear to conform to human preconceptions of how knowledge builds}. This disconnect can lead to \textit{significant misunderstandings in conversational applications}.
%
%
%
\paragraph{Social Error Patterns} \textit{InstructGPT fails to consider context, leading to lower social capability.} In particular, the clinician observed key behaviors of InstructGPT based on the Pragmatics Profile (\textbf{PP}). InstructGPT said illogical things given the surrounding context and displayed misunderstanding of directions and goals. For example, some cases are exemplified during \textbf{WC} and \textbf{USP} in Table~\ref{tab:examples}. Clinician also observed GPT provided too much information when answering questions.
Note, these contextual issues are exacerbated by an LMs limited interactive capabilities; e.g., inability to use non-verbal aspects of language and initiate. We consider how these factors affect \textbf{PP} scores through \textbf{PP}$^*$ which removes these test items: the score increases considerably, but is still far from normal for humans of any age. Overall, the limited social capabilities of instruction following models ``out-of-the-box'' suggests \textit{further work is needed to adapt them to (social) conversation applications}. 
\subsection{Automated Evaluation Results}
As before, we focus in this part on InstructGPT with comparison to ChatGPT in \S~\ref{sec:chat-compare}. Performance of InstructGPT\footnote{Intended answer is extracted using the first uttered test words (2 for \textbf{WC} \texttt{large} and 1 for \textbf{Def}); this was based on clinician observation on CELF5. Human evaluation of the rule on \textbf{WC} \texttt{large} ($n=108$) also showed 100\% intent recovery.} on \textbf{WC} \texttt{large} and \textbf{Def} is provided in Figure~\ref{fig:large-scale-acc} with $p$-values from a mean test for LM age in Figure~\ref{fig:large-scale-pvals}. We provide performance of human annotators on a 1\% ($n=108$) sample of \textbf{WC} \texttt{large} in Appendix Table~\ref{tab:humscores}.
\paragraph{Overall Performance} Coarse-grained results for InstructGPT are generally consistent with the clinical evaluation results in \S~\ref{sec:clinical_results}. Accuracy, which is equivalent to the \textbf{WC}$^*$ score in Table~\ref{tab:scores}, is consistent with the clinical evaluation based on a 95\% confidence interval.\footnote{Via Hoeffding's inequality with $n=40$ examples tested in \textbf{WC}$^*$, the two-sided interval has lower bound of 39\%.} It is notable that \textbf{WC} \texttt{large} may be more difficult, as exhibited by human disagreements (see Table~\ref{tab:humscores}). Overall, the general takeaways of the clinical exam can be confirmed in these coarse-grained results. For example, InstructGPT appears to succeed at the recollection task \textbf{Def}, which only requires recalling a definition, and perform worse at the inference task \textbf{WC} \texttt{large}. Also, GPT shows a spike in performance when word pair AoA is 19 (exactly), demonstrating unnatural word acquisition compared to human age standards.
\paragraph{Automated Determination of LM Age} Based on $p$-values in Figure~\ref{fig:large-scale-pvals}, we determine InstructGPT to align with ages 9- or 11-and-under for \textbf{WC} \texttt{large}, depending on whether $\mathbb{G}_a$ contains questions with word pair AoA exactly $a$ or $\leq a$, respectively. This can be seen by excluding all ages where the means test rejects the null that GPT aligns with age group $a$ (i.e., dipping below red line of significance). When word pair AoA is exactly 19, the means test succeeds in identifying the aforementioned "unnatural" spike in performance by correctly failing to reject the null. Overall, the means test is consistent with the clinical evaluation.
\paragraph{Automated Analysis of Errors} 
In Appendix Figure~\ref{fig:chi}, we visualize the influential factors on language errors discussed in \S~\ref{sec:error_analysis} and determine each has statistically significant association with the errors of InstructGPT. We also consider 6 hypotheses about these factors which were formulated through discussions with the evaluating clinician. Details are given in Appendix~\ref{sec:hypoth_details}. Hypotheses are tested with an LPM (see Appendix~\ref{sec:tools}), and results in Figure~\ref{fig:coeffs} confirm observations from the CELF5 exam (\S~\ref{sec:clinical_results}). We report each hypothesis and corresponding effect size $\Delta$ (increase in \% error) below:
\begin{itemize}[leftmargin=*, nolistsep]
\item \textbf{H1}: \textit{InstructGPT has more trouble when target pairs include adverbs or adjectives ($\Delta=3.5$).}
\item \textbf{H2}: \textit{InstructGPT has more trouble when the associated pair do not share POS ($\Delta=3.1$).}
\item \textbf{H3}: \textit{InstructGPT has more trouble with particular relation types ($\Delta=11$).}
\item \textbf{H4}: \textit{InstructGPT has more trouble with morphologically complex words ($\Delta=2.3$).}
\item \textbf{H5}: \textit{GPT does worse when it explains ($\Delta=6.2$).}
\item \textbf{H6}: \textit{InstructGPT has more trouble as word pair AoA increases ($\Delta=0.5$; i.e., 5\% from 9 to 19).}
\end{itemize}
\subsection{Comparison of Results with More Models}
\label{sec:chat-compare}
We focus on a comparison between InstructGPT and ChatGPT (\texttt{gpt-3.5-turbo-0301}) first, looking at both clinical and automated results. Then, we study age ranges and accuracy on \textbf{WC} \texttt{large} for a wider array of models, including newly released open-source models and more recent versions of ChatGPT (\texttt{gpt-3.5-turbo-1106}).
\begin{figure}
    \centering
    \includegraphics[width=\linewidth]{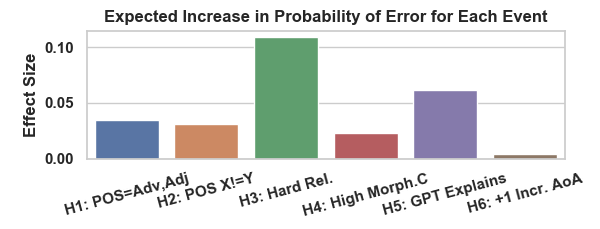}
    \caption{\small Expected increase in probability of InstructGPT error on \textbf{WC} \texttt{large} for different categories of word pairs. LPM estimates are significant at confidence 99\% (with Bonferroni correction) except \textbf{H4}. Estimates are near true effect size for large samples (see Appendix~\ref{sec:tools}).}
    \label{fig:coeffs}
\end{figure}
\begin{figure*}
    \centering
    \includegraphics[width=\textwidth]{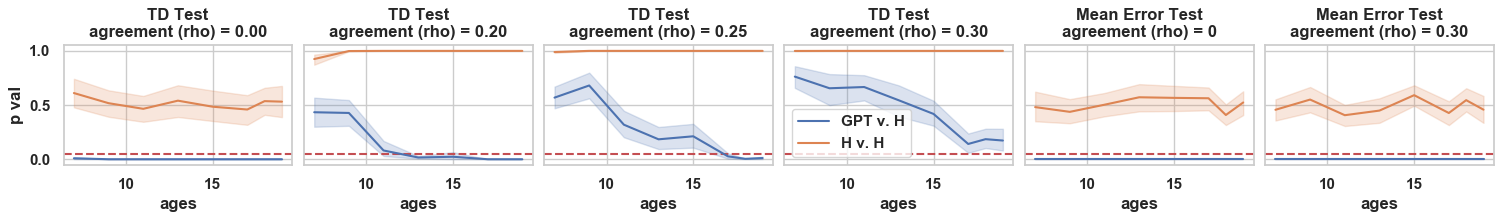}
    \caption{\small Bounds on $p$-values for $\mathtt{TD}$ and Means test. Red dotted line is significance level 0.05. $\rho$ is proportion agreement.}
    \label{fig:pvals}
\end{figure*}
\paragraph{ChatGPT Clinical Results} While focus is on InstructGPT, we also explored performance of a chat-based model (ChatGPT; \texttt{gpt-3.5-turbo-0301}) on CELF5. We focused on subtests \textbf{WC}, \textbf{USP}, and \textbf{PP}. These tests target aspects of inference and social language use (among other things) for which InstructGPT was poorly aligned with adult age groups. Findings (Table~\ref{tab:scores}) indicate ChatGPT improves upon inference about word meanings with 23\%-48\% higher scores on \textbf{WC} and \textbf{WC}$^*$ compared to InstructGPT. ChatGPT also improved upon the \textbf{PP} subtest by 9\%. Albeit, this score still aligns poorly with the pragmatics skills of adult humans. According to clinician notes, ChatGPT's safety features and limited chat medium (turn-based text) still severely limits its pragmatic abilities on CELF5. \textit{It tends to avoid providing subjective opinions (even when asked), is incapable of many non-verbal aspects of social language, and does not initiate (e.g., ask questions).}
\paragraph{ChatGPT Automated Results} We also conduct a full automated analysis on ChatGPT. The automated Mean test for LM demographic alignment shows ChatGPT aligns with ages 15-and-under when AoA = $a$ on \textbf{WC} \texttt{large}, which again agrees with the CELF5 clinical examination. In testing, the human correctness parameter $\mu_a$ for the Mean test was increased to make the Mean test more sensitive (i.e., making a more strict/difficult test), but this was still within bounds on $\mu_a$ specified by \citet{dale1976living}. The impact of changing $\mu_a$ does speak to the need for careful demographic selection, since small differences in human populations can change LM alignment. For the analysis of errors, \textbf{H1}-\textbf{H6} are consistent with results for InstructGPT, except for \textbf{H3}: ChatGPT actually does \textit{better} when it explains, whereas InstructGPT does worse. Overall, these results echo the clinician observations that \textit{ChatGPT has somewhat improved skill making new inferences about word meanings}.
\paragraph{WC \texttt{large} Test with More Models} Automated tests provide a quick and convenient tool to quantify progress in more recent model releases. We studied three 7B parameter open-source models: Llama-2-chat \citep{touvron2023llama}, Zephyr-$\beta$ \citep{tunstall2023zephyr}, and Mistral Instruct v0.2 \citep{jiang2023mistral}. For each, we use 4 bit quantization \citep{dettmers2022llmint8}. We also studied newer versions of ChatGPT (\texttt{gpt-3.5-turbo-1106}). Results in Table~\ref{tab:many_models} show open-source models tend to perform worse than InstructGPT with only Mistral Instruct v0.2 proving to be a competitive rival. Albeit, the Mistral model is still outperformed by ChatGPT. Ultimately, from these preliminary results, \textit{we expect many takeaways for smaller (7B) open-source models to be consistent with our findings on InstructGPT}; e.g., they demonstrate poor inferences about word meanings compared to human adults. As for the newest version of ChatGPT (\texttt{gpt-3.5-turbo-1106}), this model offers comparable performance to its predecessor on \textbf{WC} \texttt{large} in terms of estimated age alignment. Some degradation on words with lower AoA is observed, but this is still consistent with human populations, and moreover, is complemented by increased performance on words with higher AoA. We leave investigation of larger open-source models to future work, but expect these to narrow the gap between closed-source and open-source technology.
\paragraph{WC \texttt{large} with In-Context Learning} We also explored the impact of in-context learning (ICL) using either 3 or 10 randomly selected demonstrations. We only tested this for the newest version of ChatGPT (1106). Importantly, providing demonstrations violates aspects of the CELF5 protocol, since students are not given examples before each question. Moreover, it violates the assumptions of our statistical test, since $\mu_a$ is estimated from data on human decisions \textit{without} demonstrations. Thus, \textit{it is unclear to what extent the provided age range estimates for ICL are valid}, and we mark them with an asterisk $*$. Indeed, the issue of validity may also explain the inconsistency in age estimates for in-context learning. In any case, interpreting accuracy alone, it is easy to see that ICL tends to \textit{hurt} model performance across words with varied AoA. ICL was only beneficial with 10 examples and AoA = 19. These results may call into question the benefits of ICL when making novel inferences about semantics, i.e., echoing the discussion of what ICL really ``learns'' in recent literature \citep{min-etal-2022-rethinking, chan2022data, pan-etal-2023-context}. More thorough study of ICL, including more advanced approaches and valid statistical tests, is needed to provide confident conclusions. We leave this to future work.
\subsection{Simulated Results with $\mathbf{TD}$ Test for Age}
In the last section, we used the Means test for LM age because we did not have access to sample human question outcomes from different age groups and can only estimate the test parameter $\mu_a$. Next, we simulate data to show the benefit of the $\mathbf{TD}$ test when access to human outcomes is available.
\paragraph{Setup} Figure~\ref{fig:pvals} shows results applying tests to LM and human samples \texttt{GPT v.H} as well as two (same age) human samples \texttt{H v.H}. Ideally, a test should fail to reject the null for all \texttt{H v.H} experiments and be sensitive for \texttt{GPT v.H} experiments, rejecting the null when appropriate. To conduct tests and study variation, we require multiple human samples. Since we only have one (used to define \textbf{WC} \texttt{large}), we simulate human test performance with a random variable $H_i$ defined:
\begin{equation}\label{eqn:sim}\small
H_i = \begin{cases}
h(\hat{D}_i) & \text{with prob.} \ \rho, \\
\text{Bernoulli}\Big (\frac{\mu - \rho \mathbf{E}[h(\hat{D}_i)]}{1 - \rho} \Big ) & \text{else}
\end{cases}
\end{equation}
So, we have $\mathbf{Pr}(H_i = 1) = \mu$ regardless, and $\rho$ controls the extent to which the model \texttt{LM} and the sampled human agree. For all experiments in Figure~\ref{fig:pvals}, we conduct 25 trials. $H_i$ is simulated using Eq.~\eqref{eqn:sim}, $h(\hat{D}_i)$ is given by GPT performance on \textbf{WC} \texttt{large}, and questions for age $a$ comprise all questions whose AoA is less than or equal to $a$. We estimate $\mu$ and $\gamma$ from data.\footnote{$\mu$ is lower bound of a 95\% Hoeffding interval around the acc. in Table~\ref{tab:humscores}; $\gamma$ is  disagreement across sim. samples of $H_i$.}
%
\paragraph{Failure of Means Test} As the agreement parameter $\rho$ between the sampled human and the model $\texttt{LM}$ increases, tests using the $\mathbf{TD}$ statistic adapt appropriately, failing to reject at higher and higher ages. So, using $\mathbf{TD}$ allows us to account for context well. In comparison, the result of the means test is unchanged, demonstrating a benefit of using the $\mathbf{TD}$ statistic (when possible).
\section{Related Works} 
\label{sec:related}
%
\paragraph{Psycho-linguistic Study of LMs} 
Other tools derived from psychology and linguistics exist across previous work on LMs.
\citet{sahu2021comprehension} use Bloom's Taxonomy \citep{bloom1956taxonomy} to improve context in LM prompts for QA.
\citet{hovy-yang-2021-importance} develop a taxonomy of social factors to consider for LM evaluation. \citet{cong2022psycholinguistic} evaluate GPT-3 using psycholinguistic tests, and \citet{chang-bergen-2022-word} use word age-of-acquisition to study development of LM word knowledge (during training) compared to humans. 
Comparatively, $\mathtt{HumBEL}$ is the first work to directly measure the alignment of an LM with a human sub-population, providing systematic techniques for automatic and clinician-in-the-loop evaluation of demographic factors. With that said, recent psycho-linguistic benchmarks have begun using focused age ranges to track progress; e.g., \citet{van-duijn-etal-2023-theory} test LM Theory-of-Mind vs. children aged 7-10. While our novelty comes from the tools we propose (to test alignment, and moreover, do so across a spectrum of tasks or age ranges), it is notable our empirical results agree with recent work: both word semantics and Theory-of-Mind are challenging for many models when compared with human children, but new large commercial models have begun to show promise.
\paragraph{LM Evaluation and Human-Likeness} 
Evaluation strategies for generated text include metrics based on $n$-gram matching \citep{papineni2002bleu, lin2004rouge, vedantam2015cider} as well as metrics based on neural models \citep{sellam-etal-2020-bleurt, zhang2019bertscore, inan2021cosmic}.
\citet{bruni2017adversarial, ippolito2020automatic, dou2022gpt} also propose (human or model) adversaries to discriminate between human and generated text. 
Our work is most related to those works considering evaluation of human-likeness (and properties thereof). For example, our techniques target commonsense knowledge, inference, and social factors as studied in a variety of works \citep{nair-etal-2020-contextualized, kassner-schutze-2020-negated, liu-etal-2022-aligning}. Our work builds on broad goals of evaluating human-likeness, not only in the types of tasks we test, but also in the \textit{communication of the results to the practitioner}, presenting qualitative and quantitative results in terms of human demographic information. 
\paragraph{NLP Tasks} Many of the SLP tasks we consider have existing counterparts appearing in the NLP literature. For example, \textbf{USP} is a narrative QA task \cite{kocisky-etal-2018-narrativeqa} and \textbf{WC} is, in some respects, akin to word association tests used to evaluate semantic modeling of words \cite{bolukbasi2016man, caliskan2017semantics, liu-etal-2022-aligning}. Our work extends this literature by incorporating clinician-in-the-loop feedback for the design and evaluation of these tasks, and furthermore, is the first to incorporate human demographic data for comparison of LM performance to human sub-populations.
\section{Conclusion} 
\label{sec:disc}
We present $\mathtt{HumBEL}$, which evaluates demographic factors of conversation in language models by using novel clinician-in-the-loop statistical techniques. Our framework moves beyond measuring superficial coherence of language models, instead working towards a human-explainable way to test LMs for language use and context relevance \citep{clark1996using}, and to compare this language use to the human sub-populations that interact with these models. For example, our techniques provide insight on the utility of LMs for inference, information-extraction, and social applications. 

While the focus of this paper has been on conversational applications -- e.g., understanding the communication gaps that may persist between LMs and specific human populations -- a number of other applications of this framework are also realistic. For one, our tests can establish connections between human development and LMs (e.g., to build cognitive models), which may benefit diverse research communities in studying language disorders in humans. Moreover, testing alignment between LMs and human populations may be useful in evaluation of simulated worlds \citep{park_sim} to explore how well LMs play specific roles. While more interdisciplinary work is needed, we also hope our techniques can be extended to other factors, like in cross-cultural human-machine communication.

We make the code and data of our framework publicly available, so future researchers can make use of our suite of automated statistical techniques, and protocols for clinician evaluation.
\section*{Limitations}
\label{sec:limits}
First and foremost, we wish to be careful about claiming our proposed techniques ascribe an intellectual age to any AI model. It is not yet clear whether the tests for human language ability we use are an appropriate "all-in-one" assessment for artificial intelligence, especially considering the vast range of specific tasks in the literature at which artificial agents can achieve super-human performance. While the tasks we study are good indicators of general language skills in humans, connections between our framework and performance generalization of AI models on untested reasoning and social language tasks are unknown. For example, factors such as overfitting, adversarial robustness, stochasticity, and prompt sensitivity can all play a new distinct role for AI models. Thus, it is better to take care and interpret our framework as designed to investigate alignment of LM language use/skills to the language use/skills of \textit{particular} human demographic groups on \textit{particular} language tasks. As noted, there is still significant benefit to this more careful interpretation, since our framework serves to assess model fit in conversational AI with consideration of interlocutor demographics and goals. 

Second, the nature of language models produces a gap in evaluation protocols between children and these models. While we take a number of steps to alleviate these issues, there is still need for this gap to be bridged completely; i.e., so that normative age data is most accurate. Taking clinical evaluation to perceiving and embodied models is one possibility. One can also consider collecting new normative data on tasks designed for a language-only medium, or, consider using fine-grained metrics more commonly used by SLPs; e.g., preferring percentile rank among same age peers over age equivalents.

Third, we do not explicitly consider inter-annotator (i.e., inter-clinician agreement). The CELF5 exam \textit{does already} come with estimates of inter-clinician agreement on evaluations with humans, but it is possible that working with language models produces new challenges that will ultimately invalidate this estimate. Fourth, more human data is needed to test statistics like the test divergence on real world data. Finally, our work does not explore in-depth automated analyses on other problem areas of LMs such as social language; i.e., while our clinician-in-the-loop analysis does consider pragmatics, our automated analysis focuses on inference.%
\section*{Ethics Statement}
The proposed approach does not explicitly evaluate societal biases inherited by language models, so any harm or bias associated with these models should be considered separately. General methods that propose to mitigate harms can help to resolve these issues, along with careful human evaluations. 

For readers or users of our framework to gain access to test questions, they may need to purchase licenses from the company, university, or research lab that publishes and produces these tests. Our use of the CELF5 examination is consistent with our publishing agreement with Pearson, Inc.

Our human subject board approved our protocol. Human subjects participated voluntarily and were compensated according to the regulations approved by our human subject review board.
\section*{Acknowledgements}
Thanks to Sulare Telford Rose, Ph.D., CCC-SLP for her helpful comments and feedback.
%
\bibliography{anthology,custom}
\bibliographystyle{acl_natbib}
\clearpage
\appendix
\begin{figure*}
    \centering
    \includegraphics[width=.97\linewidth]{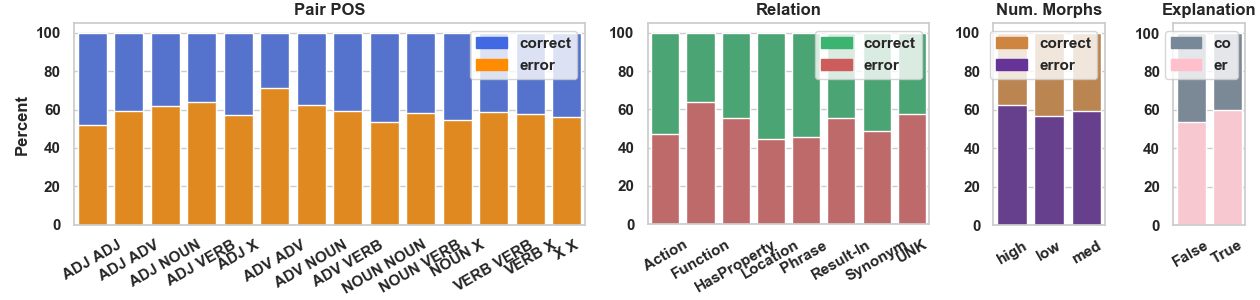}
    \caption{\small Proportion plot for features associated with InstructGPT errors on \textbf{WC} \texttt{large}. Association is significant at confidence 99\% according to $\chi^2$ test with Bonferroni correction. Infrequent categories not shown.}
    \label{fig:chi}
\end{figure*}
\section{Determination of Word AoA}
\label{sec:word_aoa}
Recall, we use a test-based age-of-acquisition dataset \citep{dale1976living, brysbaert2017test} to determine word age-of-acquisition (AoA) of 40K English words. Age is determined by U.S. K-12 grade-level and adapted to typical age equivalents (discussed later).
Word grade-level is determined via multiple-choice test in which target word definitions are provided and subjects select the target amongst multiple alternatives. A word is assigned to the earliest level at which 67-80\% of subjects answer correctly, equating to about 50-70\% of subjects "knowing" the word at this level (accounting for chance). A word's AoA is then inferred from grade-level via typical grade-to-age mapping for U.S. K-12; i.e., $\text{age} = \text{grade} + 5$. Tests were given to U.S. (Midwest) students across a range of socio-economic and racial backgrounds with each specific word-meaning administered to about 200 subjects. 
As noted, besides \textbf{WC} \texttt{large}, we also test GPT-3.5 on this multiple-choice test for matching word definitions, called \textit{Definitions} (\textbf{Def}). Alternatives are selected randomly and the prompt is: \textit{Among the words "[W]", "[X]", "[Y]", and "[Z]", the word that most means "[Defn.]" is}.
\section{Estimating Human Mean Correctness}
\label{sec:hum_mean_correct}
In experiments, we use a similar approach as \citet{dale1976living} to estimate $\mu_a$ from word AoA, accounting for guessing and subjectivity of the task. From results of \citet{dale1976living}, we make a reasonable assumption that about 50-70\% of humans at a particular age level \textit{know} a word at this age level. Unless otherwise specified, we use the lower percentage, leading to a less strict test. For a human to be correct on the \textbf{WC} task, they must both know the target words \textit{and} agree with the annotation. To compute probability for the latter, we estimate probability of agreement from Table~\ref{tab:humscores} using the upperbound of a 95\% Hoeffding interval for the reported \% disagreement (to be less strict).\footnote{Agreement is 100 less the \% disagreement. Results without the upperbound -- i.e., using exact observed disagreement-- are slightly different, but takeaways are generally consistent.} Then, assuming agreement and knowledge are independent, this means 38\% of humans aged $a$ will be correct based on knowledge. Finally, accounting for guessing using the score correction of \citet{diamond1973correction}, this means we should expect about 47\% of humans aged $a$ to answer correctly. If the higher base correctness (70\%) is assumed, $\mu_a$ is about 66\%. We assume the higher base correction in Table~\ref{tab:many_models}, for the age lower bound, and the lower base correction otherwise. Notice, our lower and upper estimates on human mean correctness are similar to those of \citet{dale1976living}, but decreased to account for the subjectivity of our task.
\begin{figure}
    \centering
    \includegraphics[width=\linewidth]{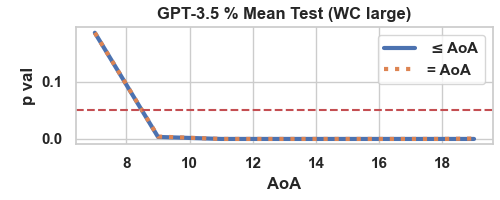}
    \caption{\small Results in Figure~\ref{fig:large-scale-pvals}, re-reported without using a Hoeffding interval to estimate disagreement. Key results (i.e., lowest age estimate) differs only by a grade level.}
    \label{fig:pvals-reprise}
\end{figure}
\section{Prompt and Parameter Sensitivity}
\label{sec:sensi_tests}
Although testing for the impact of various prompts and parameters is impractical when evaluation is done by a clinician, our automated version of the \textbf{WC} test provides a more practical alternative to explore the impact of these model choices. We test different parameter settings for nucleus sampling (i.e., $\texttt{top\_p} \in \{0.8, 0.9, 0.95\}$) and temperature scaling (i.e., $\texttt{temp} \in \{0, 0.5, 0.7, 1\}$) as well as 11 different prompts with varying aspects of the key prompt differences highlighted in Table~\ref{tab:prompts}. All in all, we test differences in InstructGPT performance on a total of 77 different prompt/parameter settings on sample of 100 examples from \textbf{WC} $\texttt{large}$. The standard deviation in the LM scores was only 3\% and a $\chi^2$ test for independence between the settings and the error rates indicates there is no statistically significant association between the settings and the error rates. That is, performance was not significantly impacted by prompt/parameter settings.
\section{Feature Extraction for Error Analysis}
\label{sec:feature_details}
\begin{enumerate}[leftmargin=*, nolistsep]
    \item \textbf{Part of Speech (POS)} While word POS is dependent on context, the explanations in the WAX dataset \citep{liu-etal-2022-wax} provide an opportunity to infer the annotator's intended POS for the word association. In particular, we can apply open-source POS parsers\footnote{We use the \texttt{spacy} package.} to the annotator explanation. This strategy assumes an explanation uses a word in the same POS as intended for the word association. In case an annotator does not use the full word pair, we use "X" for unknown. Results in Figure~\ref{fig:chi} suggest GPT-3.5 error rates can vary widely based on the pairs POS, exhibiting particular association with adverbs, adjectives, and pairs having distinct POS.
    \item \textbf{Relation} The WAX dataset also contains relation categories for word associations. Recall, the results of the clinical exam suggested particular relations are challenging for GPT-3.5 and the results in Figure~\ref{fig:chi} seem to suggest this as well; e.g., as in the clinical exam, \textit{functional} relations are hard for GPT-3.5 to identify.
    \item \textbf{Morphological Complexity} We also consider Morphological Features within the Universal Dependencies framework \citep{nivre-etal-2016-universal}, which describe semantic and grammatical properties of words. We define \textit{morphological complexity} as the total number of morphological features attached to at least one of the the words in the association. \textit{High} corresponds to more than 4 features, \textit{medium} corresponds 3-4 features, and \textit{low} corresponds to 2 or less features. Our working assumption is that the number of features is a loose indicator of the complexity of the a word's meaning and can thus introduce challenges for GPT-3.5. The results in Figure~\ref{fig:chi} do appear to confirm this hypothesis.
    \item \textbf{Explanations} Lastly, we consider if GPT-3.5 provides an (unprompted) explanation of its reasoning behind an answer. Interestingly, this occurs more times than not on the \textbf{WC} \texttt{large} dataset. While our intuition may tell us this means GPT-3.5 is more confident in the answer, the clinical evaluation actually demonstrated that GPT-3.5 often provided illogical explanations that may appear off-topic or overly complex to humans. Results in Figure~\ref{fig:chi} seem to confirm these findings, indicating that explanations typically led to worse performance at identifying associations.
\end{enumerate}
\section{Hypothesis Selection}
\label{sec:hypoth_details}
Below, we provide some details discussed with the evaluating clinician which led to the suite of hypotheses we test.
\begin{itemize}[leftmargin=*, nolistsep]
    \item \textbf{H1}: \textit{InstructGPT has more trouble when the associated pair includes an adverb or adjective.} Clinician observations indicate trouble with modifiers in CELF5 examination. This hypothesis is confirmed in Figure~\ref{fig:coeffs} where we estimate a 3.5\% increase in probability of error when at least one word in the pair is an adjective or adverb. 
    \item \textbf{H2}: \textit{InstructGPT has more trouble when the associated pair do not share POS.} Distinct POS can indicate more complex relationships across word pairs, which is a noted  problem for GPT in CELF5 evaluation. This hypothesis is confirmed with a similar effect size as \textbf{H1}.
    \item \textbf{H3}: \textit{InstructGPT has more trouble with particular relation types.} Building on the last hypothesis, we isolate "easy" word pair relations including \{\textit{action}, \textit{location}, \textit{phrase}, and \textit{synonym} \}, so the remaining "hard" word pair relations overlap with types of relations our clinician noted as difficult for GPT. Unknown relations are assumed to be hard. Results in Figure~\ref{fig:coeffs} confirm this hypothesis where we estimate a relatively large 11\% increase in error probability for "hard" relations.
    \item \textbf{H4}: \textit{InstructGPT has more trouble with morphologically complex words.} As before, assuming the complexity of a word is tied to its count of morphological features, we would expect GPT to have trouble with words having \textit{medium} or \textit{high} morphological feature count. We estimate an effect size similar to \textbf{H1} and \textbf{H2}.
    \item \textbf{H5}: \textit{InstructGPT does worse when it explains.} Clinician evaluation on the Pragmatics checklist reveals untrustworthy, illogical explanations by GPT. Testing at scale reveals GPT has more errors when it attempts to explain its reasoning with a relatively large estimated effect size of 6\%. 
    \item \textbf{H6}: \textit{InstructGPT has more trouble as the word pair AoA increases.} While we include word pair AoA in our analysis as a potential confounder for which to control, it is also interesting to see how this variable impacts the performance of GPT. We estimate a 0.5\% increase in probability of error for each unit increase in AoA; e.g., a word pair AoA of 19 would cause 5\% greater chance of error than an AoA of 9.
\end{itemize}
\section{Overview of Statistical Tools}
\label{sec:tools}
\paragraph{$\chi^2$ Test}
The $\chi^2$ test is commonly used to determine statistical association between two categorical variables \citep{freund2004john}. In our case, the two categorical variables are (1) the occurrence of a language application error by GPT and (2) one of the categorical features of the word pair discussed in \S~\ref{sec:error_analysis}. The test uses a \textit{contingency table}; i.e., a table of counts formed by letting one of the variables define the columns, the other variable define the rows, and filling each element with the number of occurrences observed for each pair of categories. Then, the test uses the statistic
\begin{equation}\small
\chi^2 = \sum\nolimits_{i = 1}^k \frac{(\text{observed}_i - \text{expected}_i)^2}{\text{expected}_i}
\end{equation}
where $k$ is the number of elements in the contingency table, $\text{observed}_i$ is the observed frequency of each element of the table, and $\text{expected}_i$ is the expected frequency under the assumption that the two categorical variables are independent (i.e., the null hypothesis). Aptly, the distribution of the statistic is asymptotically $\chi^2$ and a $p$-value can be calculated accordingly. We use a Bonferroni correction to control for multiple testing (i.e., across the multiple features we present as well as those not presented). 
\paragraph{Linear Probability Model}
Consider a $n\times 1$ vector of dependent variables $Y$ and a $n \times m$ matrix of independent variables $X$ where $n$ is the number of observations and $m$ is a number of features for each observation. In our case, $Y$ is a binary vector indicating the occurrence of a GPT language application error and $X$ is a matrix ($m=4$) with the 3 categorical features (discussed in \S~\ref{sec:error_analysis}), and the last column being the word pair AoA (\S~\ref{sec:wc_large}). With this notation, the Linear Probability Model (LPM) assumes a conditional probability model:
\begin{equation}\small
\mathbf{Pr}(Y=1 | X) = 
\begin{cases}
1, & \quad X\beta > 1 \\
0, & \quad X\beta < 0 \\
X\beta , & \quad \text{else}
\end{cases}
\end{equation}
where $\beta$ is an unknown parameter vector of implied dimension. Supposing $\mathbf{Pr}(X\beta > 1) = \mathbf{Pr}(X\beta < 0) = 0$, the LPM reduces to the assumption: $\mathbf{Pr}(Y=1 | X) = X\beta$,
in which case, the standard OLS estimate
\begin{equation}\small
\label{eqn:ols}
\hat{\beta} = (X^\mathrm{T}X)^{-1}X^\mathrm{T}Y
\end{equation}
provides a consistent estimator for the true parameter $\beta$ \citep{horrace2003new}. Techniques for heteroscedasticity (i.e., unequal variance of errors) like White's robust covariance matrix \citep{white1980heteroskedasticity} can also be used to conduct hypothesis testing for significance of the coefficient estimates \citep{horrace2003new}. We use these techniques for the coefficient estimates and statistical tests in \S~\ref{sec:results} Figure~\ref{fig:coeffs}. As before, we employ a Bonferroni correction to control for multiple testing.%
\label{sec:additional_results}
\begin{table}[]
    \centering\small
    \begin{tabular}{c|c|c||c|c}
        Hum. & A1 $\neq$ A2 & $\kappa$ & GPT & $\neq$ Hum. \\\hline
        84\% & 15\% & 0.82 & 56\% & 40\% \\
    \end{tabular}
    \caption{\small Sample ($n=108$) \textbf{WC} \texttt{large} scores of 2 annotators aged 19+ (left) and InstructGPT (right). Annotators \% disagreement and Cohen's $\kappa$ is reported. GPT avg. \% disagreement with annotators is reported. Annotators were
    prompted using 
    the same directives as GPT; i.e., \textit{which two words go together best?}}
    \label{tab:humscores}
\end{table}%
\begin{figure}
    \centering
    \includegraphics[width=\linewidth]{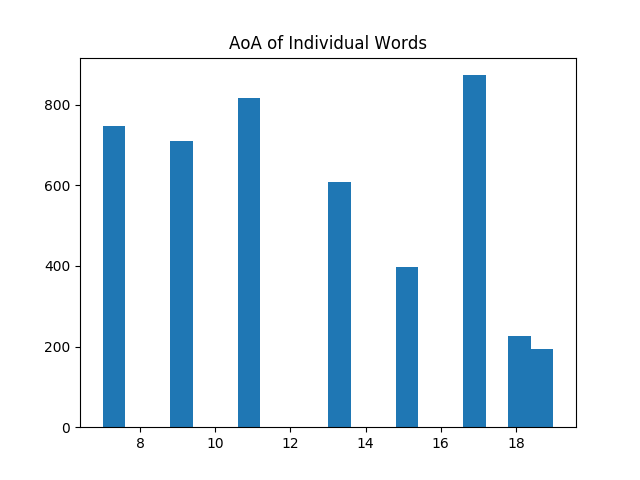}
    \caption{\small AoA of individual words from dataset of \citet{dale1976living} used to create \textbf{WC} \texttt{large}.}
    \label{fig:wordaoa}
\end{figure}
\begin{figure}
    \centering
    \includegraphics[width=\linewidth]{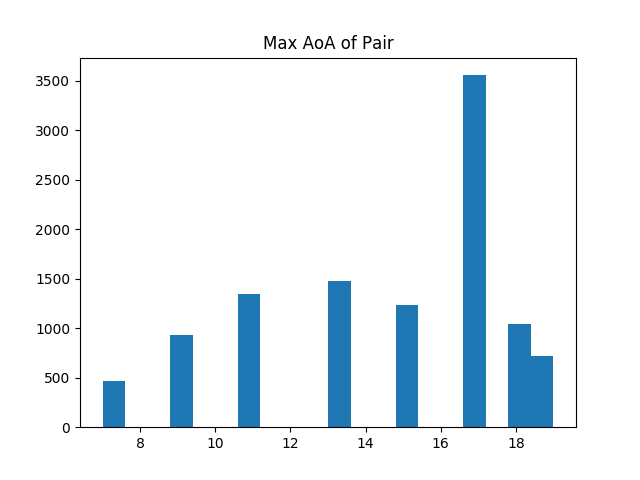}
    \caption{\small AoA of word pairs in \textbf{WC} \texttt{large}. Some expected accumulation in higher ages occurs (i.e., from taking a max).}
    \label{fig:pairaoa}
\end{figure}
\paragraph{Drawbacks of LPMs} Notably, the LPM has been criticized by some because it is a somewhat fragile model of the Bernoulli process governing $Y$ \citep{gomila2021logistic}. For example, if $X\beta > 1$ or $X\beta < 0$ are probable, the interpretation of the model is unclear. Indeed, mathematically, when the presumed model is not true (e.g., when there are data such that $X\beta > 1$) the least square estimates for the LPM coefficients in Eq.~\eqref{eqn:ols} are biased \citep{horrace2003new}. For this reason, Logistic Regression is often used instead. In our case, via standard testing procedures, one cannot refute the correctness of the LPM with data \citep{horrace2003new, battey2019linear}. Further, a logistic regression analysis led to the same takeaways as presented in the main text. Thus, we opt to show results for an LPM in the main text, since these are generally more easily interpreted (i.e., they show percent change instead of change in log odds).
\end{document}